\pdfoutput=1

\documentclass[11pt]{article}

\usepackage{ACL_Style_files/acl}

\usepackage{times}
\usepackage{latexsym}

\usepackage[T1]{fontenc}

\usepackage[utf8]{inputenc}

\usepackage{microtype}

\usepackage{inconsolata}

\usepackage{graphicx}
\usepackage{longtable}
\usepackage{supertabular}
\usepackage{caption}
\usepackage[shortlabels]{enumitem}
\usepackage{subcaption}  
\usepackage{multirow, multicol}
\usepackage{amssymb, amsfonts, amsmath, bbm, mathtools}
\usepackage{textcomp}  
\usepackage{tcolorbox}

\usepackage{xfrac, nicefrac}  
\usepackage{float}
\usepackage{stfloats}
\DeclareMathOperator*{\argmax}{arg\,max}
\usepackage[normalem]{ulem}
\useunder{\uline}{\ul}{}

\title{Enhancing Small Language Models for Cross-Lingual Generalized Zero-Shot Classification with Soft Prompt Tuning}

\author{Fred Philippy\textsuperscript{1,2}, Siwen Guo\textsuperscript{1}, Cedric Lothritz\textsuperscript{3}, Jacques Klein\textsuperscript{2}, Tegawendé F. Bissyandé\textsuperscript{2} \\ \\ \textsuperscript{1} Zortify Labs, Zortify S.A., Luxembourg \\ \textsuperscript{2} SnT, University of Luxembourg, Luxembourg \\ \textsuperscript{3} Luxembourg Institute of Science and Technology (LIST), Luxembourg \\ \texttt{\{fred, siwen\}@zortify.com} \hspace{1cm} \texttt{cedric.lothritz@list.lu} \\ \texttt{\{tegawende.bissyande, jacques.klein\}@uni.lu}}

\usepackage{ifthen}
\usepackage{amssymb}

\newboolean{showcomments}
\setboolean{showcomments}{true}
\ifthenelse{\boolean{showcomments}}
 { \newcommand{\mynote}[2]{
      \fbox{\bfseries\sffamily\scriptsize#1}
        {\small$\blacktriangleright$\textsf{\emph{#2}}$\blacktriangleleft$}}}
        { \newcommand{\mynote}[2]{}}

\usepackage{tcolorbox}

\begin{document}
\maketitle
\begin{abstract}
In NLP, Zero-Shot Classification (ZSC) has become essential for enabling models to classify text into categories unseen during training, particularly in low-resource languages and domains where labeled data is scarce. While pretrained language models (PLMs) have shown promise in ZSC, they often rely on large training datasets or external knowledge, limiting their applicability in multilingual and low-resource scenarios.
Recent approaches leveraging natural language prompts reduce the dependence on large training datasets but struggle to effectively incorporate available labeled data from related classification tasks, especially when these datasets originate from different languages or distributions. Moreover, existing prompt-based methods typically rely on manually crafted prompts in a specific language, limiting their adaptability and effectiveness in cross-lingual settings.
To address these challenges, we introduce RoSPrompt, a lightweight and data-efficient approach for training soft prompts that enhance cross-lingual ZSC while ensuring robust generalization across data distribution shifts. RoSPrompt is designed for small multilingual PLMs, enabling them to leverage high-resource languages to improve performance in low-resource settings without requiring extensive fine-tuning or high computational costs. We evaluate our approach on multiple multilingual PLMs across datasets covering 106 languages, demonstrating strong cross-lingual transfer performance and robust generalization capabilities over unseen classes.


\end{abstract}
\section{Introduction}

Zero-Shot Classification (ZSC) is a task in NLP where a model classifies inputs into classes that it has not seen during training. 
This task is crucial in real-world scenarios where some classes are underrepresented with little or no labeled data.
Traditionally, two approaches have dominated the landscape: entailment-based and similarity-based approaches. Entailment-based approaches \citep{yin_benchmarking_2019} focus on understanding relationships between sentences, particularly determining the level of entailment between the document and the potential class labels. This method requires the model to have a deep understanding of language structure and logic. On the other hand, similarity-based approaches focus on computing the similarity between the input and labels of each class, even if the model has never encountered them during training. This method often relies on embeddings or vector representations of text, allowing the model to make inferences based on how closely the input aligns with class descriptors \citep{schopf_evaluating_2023}.

However, these methods face inherent drawbacks, as they depend on Natural Language Inference or Semantic Text Similarity datasets that require considerable effort to develop and are susceptible to potential biases \citep{pavlick_inherent_2019, kalouli_curing_2023}. In light of this, and with the acknowledgment of the extensive knowledge embedded in general pre-trained language models (PLMs) and the potential to extract it, a novel paradigm has arisen: prompting. Prompting reformulates a task as a cloze-style task using a natural language prompt, retrieves the model's masked or next token prediction, and maps it to the right class via a verbalizer, while requiring little to no training data.
Nevertheless, traditional prompting methods are hindered not only by manual effort and inherent biases of the individuals creating the prompts and verbalizers, but also by other factors such as the order of examples in the prompt during in-context learning \citep{zhao_calibrate_2021, lu_fantastically_2022}.

To address this, \citet{shin_autoprompt_2020} developed an automated system for generating prompts and verbalizers using a limited number of training samples. 
Furthermore, \citet{hu_knowledgeable_2022} introduced a technique that eliminates the need for training data by automatically creating a verbalizer using an external knowledge base. Motivated by the goal of eliminating the need for any additional data, \citet{zhao_pre-trained_2023} proposed a method that forms a verbalizer using only the PLM's embedding space, without requiring any training data or external knowledge base.
This approach, while efficient and effective in various ZSC tasks, shares a limitation with the methods of \citet{shin_autoprompt_2020} and \citet{hu_knowledgeable_2022}: it relies on language-specific prompts which introduce a language bias, making the method less effective in multilingual scenarios.
Moreover, despite the high efficiency and appeal of methods that operate without existing data, their inability to leverage even a minimal amount of available data from a similar classification task in a high-resource language, can be seen as a significant limitation in our data-abundant world.

To address these shortcomings, we suggest to transform the language-specific hard prompts into trainable soft prompts \citep{lester_power_2021}, which can then be fine-tuned. However, directly adopting the conventional soft prompt tuning (SPT) setup leads to overfitting on the seen classes (\S \ref{sec:GZSL}), therefore, does not generalize under data distribution shifts. In response to this constraint, we introduce \textit{\textbf{Ro}bust \textbf{S}oft \textbf{Prompts}} (\texttt{\textbf{RoSPrompt}}), a novel method for cross-lingual zero-shot topic classification through few-shot SPT, which exhibits robust out-of-distribution generalization and strong cross-lingual transfer performance. \texttt{RoSPrompt} not only retains the efficiency and effectiveness of leveraging the knowledge of PLMs but also enhances it by incorporating small sets of existing data. By doing so, we aim to broaden the applicability of ZSC in a multilingual context, ensuring more accurate topic classification across diverse languages and datasets.

Specifically, our approach
\begin{enumerate}[(a)]
\item enables the training of soft prompts, which are better suited for ZSC tasks compared to hand-crafted, natural language hard prompts.
\item shows strong cross-lingual transfer performance after few-shot fine-tuning in English, with soft prompts significantly improving accuracy compared to hard prompts.
\item displays significant robustness against data distribution shifts, enabling the fine-tuning of the prompt on any available topic classification data for subsequent use in diverse topic classification tasks.
\item exhibits computational efficiency, as fewer than 1\% of parameters are fine-tuned in comparison to full-model fine-tuning.
\end{enumerate}

To showcase the efficacy of our proposed approach, we conduct a comprehensive evaluation using three distinct types of multilingual language models (encoder-only, decoder-only, and encoder-decoder) and three diverse datasets, encompassing 106 languages, thereby highlighting the versatility and applicability of our method in cross-lingual scenarios.

\section{Background} \label{sec:related}
\paragraph{Soft Prompt Tuning (SPT)}
Our approach is based on SPT \citep{lester_power_2021}, extending it specifically for cross-lingual zero-shot topic classification. SPT appends tunable vectors (soft prompts) to the input of a PLM, training only the soft prompts while keeping the original model weights frozen. This method demonstrates efficacy in various downstream tasks, providing a balance between model performance and resource efficiency, and is particularly effective for cross-lingual transfer \citep{philippy_soft_2024}.

Given an input sequence $\mathbf{x}$ and the set of $C$ potential classes $\mathcal{C}$, we define the two main components of SPT:
\begin{itemize}
    \item A \textbf{soft prompt} $\mathbf{p}$ that is appended to $\mathbf{x}$ in order to obtain $\mathbf{x}'=[\mathbf{x};\mathbf{p}]$, where $[\cdot;\cdot]$ is the concatenation function.
    \item A \textbf{verbalizer} $v:\mathcal{T}\rightarrow \mathcal{C}$ which maps the token predicted by the model to the respective class.
    $\mathcal{T}=\{t_1,\ldots,t_C\}$ is a subset of the model's vocabulary $\mathcal{V}$ and the token $t_c$ "describes" the class $c$.
\end{itemize}

If we denote the function performed by the model as $f$, with its parameters $\theta$ (which are frozen during SPT), the logits over the vocabulary $\mathcal{V}$ for the next token in the sequence are given by:
$$
f_\theta(\mathbf{x}') = \{z_1,\ldots,z_{|\mathcal{V}|}\}
$$
The predicted class will then be
$$
\hat{y} =  \argmax_{c \in \mathcal{C}} z_{t_c}
$$

\paragraph{Nonparametric Prompting (NPPrompt)}
\citet{zhao_pre-trained_2023} demonstrated that PLMs possess significant innate capabilities for ZSC, even without task-specific fine-tuning. Their technique, NPPrompt, involves adding a natural language prompt to the input example, prompting the model to fill in the \texttt{[MASK]} for BERT-based models, or predict the next token for autoregressive and Seq2Seq models, which are then used for the final classification of the sample. Nevertheless, their strategy is primarily designed for English, as the prompts employed are in English. Applying their method to additional languages would necessitate the engineering of new prompts specific to those languages. Furthermore, despite the appeal of their zero-shot framework, particularly when there is a lack of fine-tuning data, it falls short by not accommodating the use of existing labeled data when it is available. Therefore, we suggest to extend their method by transforming the natural language prompt into a trainable soft prompt \citep{lester_power_2021}, enabling its training through any available topic classification data in the source language for subsequent zero-shot topic classification in any target language.
\section{\texttt{RoSPrompt}}
\begin{figure}[H]
    \centering
    \includegraphics[width=0.48\textwidth]{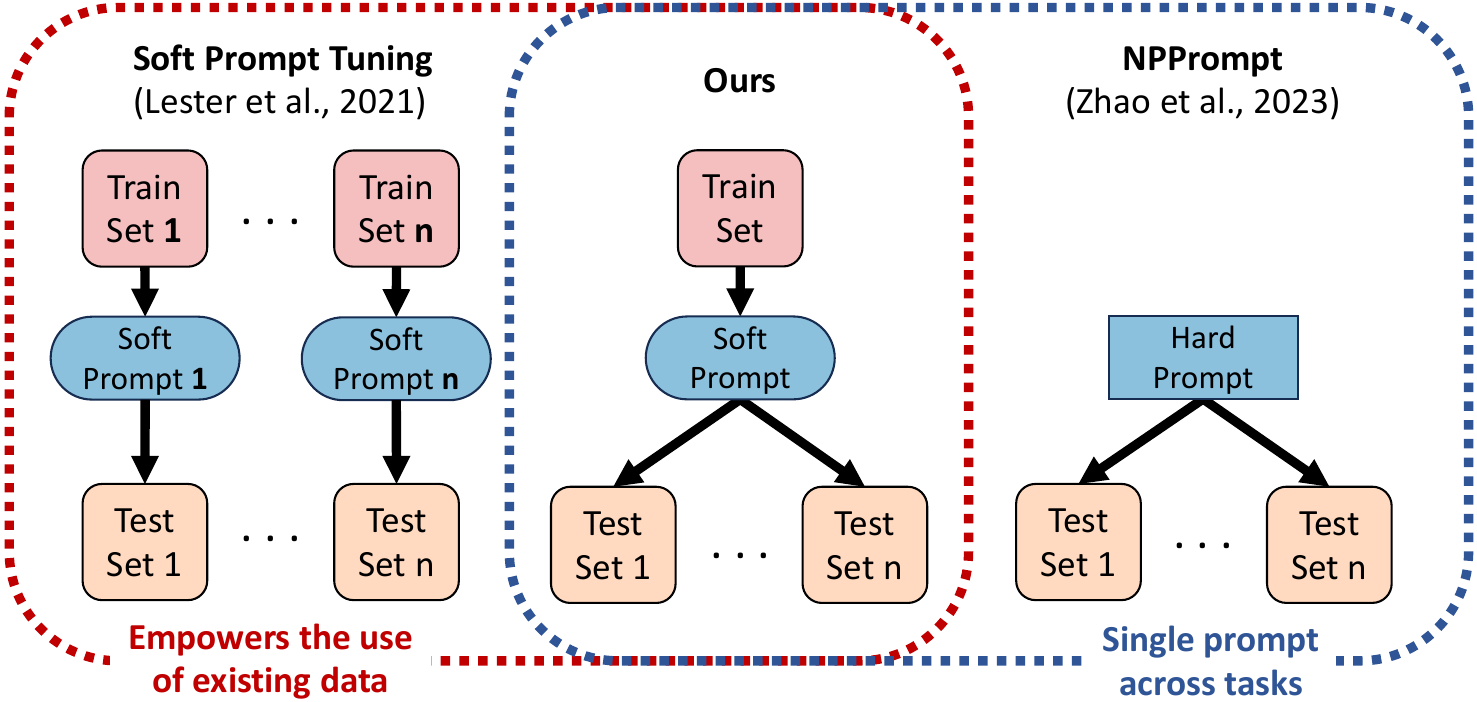}
    \caption{Conventional SPT \citep{lester_power_2021}, while effective in leveraging existing data, requires distinct training for each topic classification task. Conversely, NPPrompt \citep{zhao_pre-trained_2023} offers versatility with a single natural language prompt for various tasks but lacks data leverage. Our method combines the strengths of both methods, enabling data utilization with a single soft prompt applicable across diverse topic classification tasks, while effectively overcoming the drawbacks of both methods. }
    \label{fig:method_comparison}
\end{figure}

We describe our technique as a hybrid of SPT \citep{lester_power_2021} and NPPrompt \citep{zhao_pre-trained_2023}. SPT excels in data efficiency but is sensitive to data distribution shifts, needing unique prompts for each topic classification dataset. On the other hand, NPPrompt uses one prompt for various data distributions but fails to leverage existing data. Our strategy combines their strengths, using a single, robust soft prompt for different data distributions and enhancing data utilization (Figure \ref{fig:method_comparison}).

\begin{figure*}[h]
    \centering
    \includegraphics[width=\textwidth]{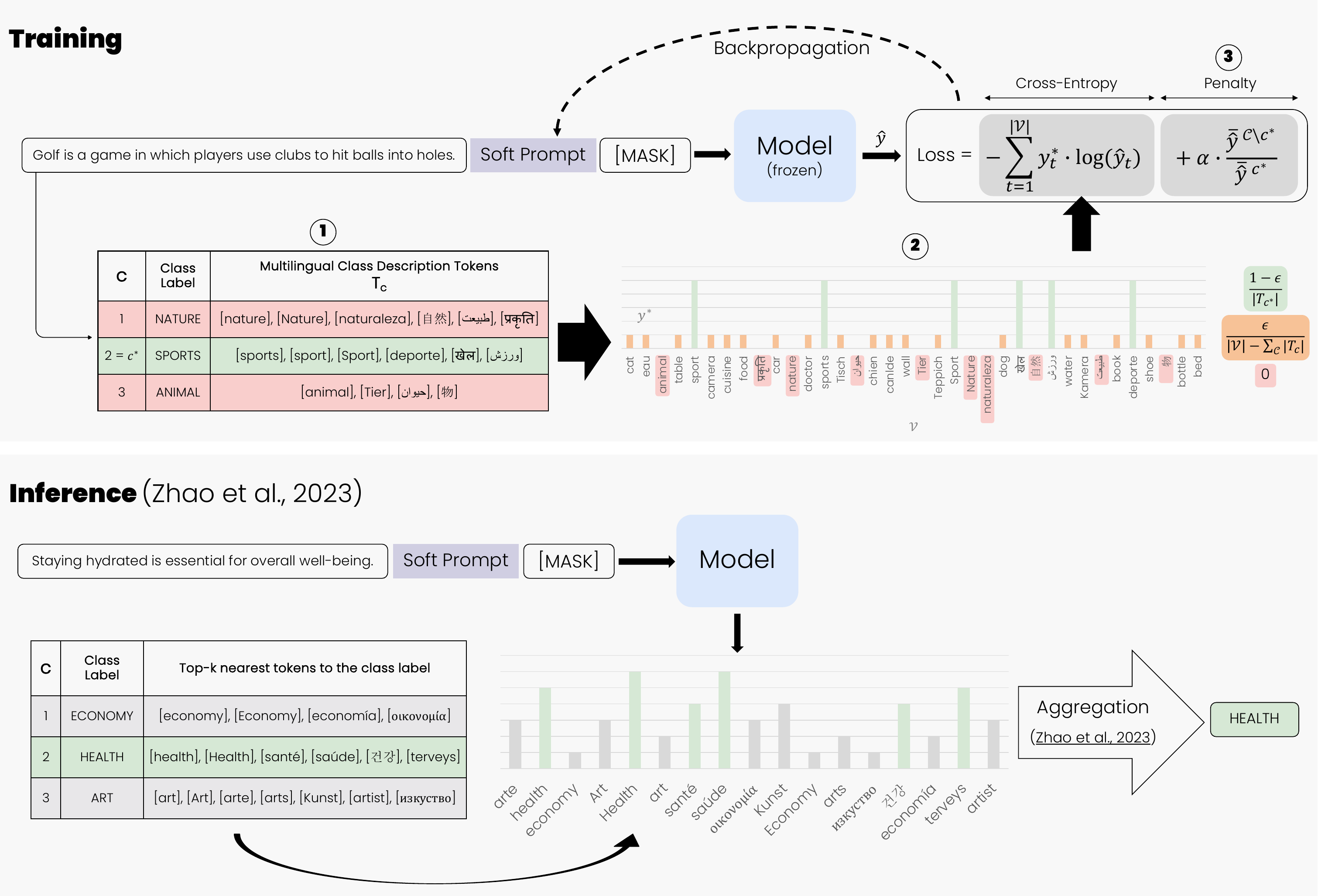}
    \caption{Visual representation of \textbf{\texttt{RoSPrompt}}. During \underline{training}, each class is categorized by a \textbf{multilingual set of label tokens} (\textcircled{\textbf{\scriptsize{1}}}). We apply \textbf{contrastive label smoothing} (\textcircled{\textbf{\scriptsize{2}}}) to the probability distribution across the entire vocabulary. To further deter overfitting, we integrate a \textbf{custom penalty} (\textcircled{\textbf{\scriptsize{3}}}) into the loss function. During \underline{inference}, we retrieve the logits predicted by the model and use the aggregation technique proposed by \citet{zhao_pre-trained_2023} to make the final prediction.}
    \label{fig:our_method}
\end{figure*}

Figure \ref{fig:our_method} provides a graphical illustration of our approach. The novelty of our method is most apparent in the training phase (\S \ref{sec:our_method_training}), which involves three main components: \textbf{1)} the application of a multilingual verbalizer; \textbf{2)} the use of contrastive label smoothing; \textbf{3)} the adoption of a custom loss function penalty. For the inference phase of our method, we adopt the technique proposed by \citet{zhao_pre-trained_2023}, aligning seamlessly with our goals.

\subsection{Training} \label{sec:our_method_training}
Below, we detail the three main components of our training approach.

\paragraph{1) Multilingual Class Description Tokens}
As mentioned before, in the standard methodology of SPT, a class $c$ is characterized, via the verbalizer, by a single token $t_c$ from the vocabulary $\mathcal{V}$. 
However, this single token might not fully capture the essence of the respective class. 
Moreover, it is confined to one language, leading to potential inconsistencies in multilingual settings, where the sample and the verbalizer token may be in different languages.

Therefore we propose, during training, to extend the single verbalizer token  $t_c$ to a multilingual set of verbalizer tokens $T_c = \{t^{(1)}_c, t^{(2)}_c, \ldots \}$. These augmented verbalizer tokens could be additional descriptive tokens, such as synonyms or translations of the original label token.

Our method does not mandate a uniform number of verbalizer tokens across different classes, and the manual labor involved in generating these labels is a one-time effort only required for fine-tuning the soft prompt.

\paragraph{2) Contrastive Label Smoothing}
Conventionally, when pre-training large language models, using self-supervised tasks such as the masked language modeling or next-token prediction objective, a single token from the vocabulary is considered to be the gold truth.

Mathematically, given a token vocabulary $\mathcal{V}$, $y=\left[y_1,\ldots,y_{|\mathcal{V}|}\right]$  represents the "true" masked or next token in one-hot encoded form. When using a "hard" probability distribution, if $t^*$ is the "true" token, $\forall t \in \mathcal {V}$,

$$y_t = 1 \times \mathbbm{1}_{\{t=t^*\}}$$

for the cross-entropy loss defined as

$$
CE(\hat{y},y) = - \sum_{t=1}^{|\mathcal{V}|} y_t \times \log\left(\hat{y}_t\right)
$$
where $\hat{y}$ represents the probabilities predicted by the model.

In other words, this standard method assigns a probability of 1 to the true token and 0 to the others, which might lead to overfitting as the model becomes overly confident in certain predictions. A strategy to resolve this is label smoothing \citep{szegedy_rethinking_2016}, a regularization technique penalizing models for over-confident predictions and thereby mitigating overfitting. Label smoothing achieves this by shifting from a "hard" probability distribution, where only the true token gets a non-zero probability, to a "soft" distribution, where small probabilities are allocated to all or some vocabulary tokens, and the probability for the true token is reduced.

Our method employs a modified form of conventional label smoothing, which we refer to as \textit{contrastive label smoothing}. 
This variation is designed to handle multiple "true" tokens for each class. 
Additionally, it not only prevents overconfident predictions by the model but also penalizes it for consistently favoring class label tokens over those without a class assignment. 
We argue that this approach leads to improved generalization over unseen classes in ZSC setups. 

If $\mathcal{C}$ represents the potential classes of the training data, we denote $\left(T_c\right)_{c\in \mathcal{C}}$ as the label class token collections for each class, where $T_c$ is the collection of verbalizer tokens of class $c$. If a sample belongs to class $c^*$ we distribute the probabilities across the vocabulary, $\forall t \in \mathcal {V}$, as follows:
$$
y_t = \left\{\begin{array}{cc}
     \frac{1-\epsilon}{|T_{c^*}|} & \text{if } t \in T_{c^*} \vspace{0.2cm} \\
     \frac{\epsilon}{|\mathcal{V}|-\sum_{c \in \mathcal{C}} |T_c|} & \text{if } t \notin \displaystyle \bigcup_{c \in \mathcal{C}} T_c \vspace{0.2cm} \\
     0 & \text{otherwise}
\end{array}   \right.
$$
In other words we uniformly distribute a collective probability of $1-\epsilon$ over the label tokens of the true class, i.e. $T_{c^*}$, and the remaining probability $\epsilon$ over all other tokens in the vocabulary \textbf{except} the label tokens of other classes.

\paragraph{3) Penalty}
In order to further penalize the soft prompt for overfitting on the seen classes during training, we additionally add a penalty to the cross-entropy loss function. 
We define
$$
\boldsymbol{\bar{\hat{y}}^{\hspace{0.05cm} \mathcal{C}\backslash c^*}} = \frac{\displaystyle \sum_{c \in \mathcal{C} \backslash c^*} \hspace{0.05cm} \sum_{t \in T_c} \hat{y}_t}{\displaystyle \sum_{c \in \mathcal{C} \backslash c^*}|T_c|}
\quad \textbf{and} \quad
\boldsymbol{\bar{\hat{y}}^{c^*}} = \frac{\displaystyle \sum_{t \in T_{c^*}} \hat{y}_t}{\displaystyle |T_{c^*}|}
$$
as the average predicted probabilities for all verbalizer tokens across all classes except the true class $c^*$, and for all verbalizer tokens within the class $c^*$, respectively.

With these definitions, we express the penalty $\Omega$ as:
$$
\Omega\left(\hat{y}\right) = \frac{\bar{\hat{y}}^{\hspace{0.05cm} \mathcal{C}\backslash c^*}}{\bar{\hat{y}}^{c^*}}
$$

This penalty simply expresses the ratio of the average predicted probabilities for the true class tokens and the class tokens for all other potential classes.

Hence, the final loss function used in our approach becomes
$$
L(\hat{y},y) = CE(\hat{y},y) + \alpha \times \Omega\left(\hat{y}\right)
$$
where $\alpha$ is the coefficient that controls the influence of the penalty.

\subsection{Inference} \label{sec:our_method_inference}
During inference we use the methodology proposed by \citet{zhao_pre-trained_2023}.

The verbalizer tokens get automatically chosen by selecting the the top-k nearest tokens in the embedding space to each original English class label $t_c$. More specifically, for a given class $c$, its verbalizer tokens are given by
$$
T_c = \underset{t \in \mathcal{V}}{\text{Top-k}} \left\{ S(\text{emb}(t), \text{emb}(t_c)) \right\}
$$
where $S(\cdot)$ is the cosine similarity function.

For a given input document $x$, the aggregated prediction score for class $c$, based on the model's output logits for the next or MASK token, $\hat{y}$, is given by
$$
Q(c | x) = \sum_{t \in T_c} w(t, t_c) \cdot \hat{y}_t
$$

where the weight of each token in the verbalizer for a given class $c$ is given by
$$
w(t, t_c) = \frac{\exp(S(\text{emb}(t), \text{emb}(t_c)))}{\sum_{j \in T_c} \exp(S(\text{emb}(j), \text{emb}(t_c)))}
$$

The final predicted class is then given by
$$
\hat{c} = \argmax_{c \in \mathcal{C}} Q(c | x)
$$
This selects the class with the highest aggregated prediction probability.
\section{Experimental Setup}
We provide a general description of the datasets for training and evaluation, along with the models used in our experiments. Further specific details about the experimental setup can be found in Appendix \ref{app:technical_details}.

\subsection{Datasets}
For our experiments, a general English document classification dataset serves as the source data for training the soft prompts. We then evaluate these prompts on three diverse multilingual datasets, each with its own set of classes.
\subsubsection{Training}
As training data we use the English \textbf{\texttt{DBPedia14}} dataset, an ontology classification dataset, compiled from Wikipedia's most frequently used infoboxes and containing 14 distinct classes. Every class includes 40.000 samples for training and 5.000 samples for testing.

\subsubsection{Evaluation}
For evaluation we use 3 distinct multilingual topic classification datasets. Further details being provided in Appendix A.3.

\vspace{0.1cm}

\noindent \textbf{\texttt{MLSUM}} \citep{scialom_mlsum_2020}, a multilingual news summarization dataset. We classify articles based on their summaries, using six main categories per language, although the exact categories differ slightly across languages.

\vspace{0.1cm}

\noindent \textbf{\texttt{MTOP}} \citep{li_mtop_2021}, a multilingual utterance classification dataset, featuring 11 different domains and covering 6 languages.

\vspace{0.1cm}

\noindent \textbf{\texttt{SIB-200}} \citep{adelani_sib-200_2024}, a multilingual topic classification dataset featuring 7 categories and covering more than 200 languages.

\vspace{0.1cm}

We focus on using MTOP and MLSUM to test the robustness of our method under distribution shifts, but since they are limited to high-resource languages, we leverage SIB-200 to assess cross-lingual transfer to low-resource languages, thanks to its broader language coverage

\subsection{Models}
We evaluate our method on three distinct models, each one based on a different architecture:

\vspace{0.1cm}

\noindent \texttt{\textbf{XGLM}} \citep{lin_few-shot_2022}, a \textit{decoder-only} model supporting 30 different languages.

\vspace{0.1cm}

\noindent \texttt{\textbf{mT0}} \citep{muennighoff_crosslingual_2023}, an \textit{encoder-decoder} model supporting 101 languages, which is a multi-task fine-tuned version of the mT5 model \citep{xue_mt5_2021}.

\vspace{0.1cm}

\noindent \texttt{\textbf{XLM-R}} \citep{conneau_unsupervised_2020}, an \textit{encoder-only} model, supporting 100 languages.

\vspace{0.1cm}

More specifically, we use the \texttt{XGLM-564M}, \texttt{mT0-base} and \texttt{XLM-RoBERTa-large} variants. We describe them in more detail in Appendix A.4.

\subsection{Baselines}
We evaluate RoSPrompt against different baselines:

\vspace{0.1cm}

\noindent \textbf{NPPrompt} \citep{zhao_pre-trained_2023}, previously described in Section \ref{sec:related}, using the English hard prompt "In this sentence, the topic is about [MASK]".

\vspace{0.1cm} 

\noindent \textbf{NPPrompt-t}, a variant of NPPrompt where the English prompt is translated into the target language for inference.\footnote{Languages unsupported by Google Translate or with syntax that does not place the [MASK] token at the end are excluded. In Table \ref{tab:sib-200}, English prompt performance is used for reporting.}

\vspace{0.1cm}

\noindent \textbf{SPT} \citep{lester_power_2021}, previously described in Section \ref{sec:related},  where a soft prompt is fine-tuned on English samples using standard SPT practices and then used with NPPrompt during inference.

\vspace{0.1cm}

\noindent \textbf{Zero-Shot Prompting}, where we evaluate generative LLMs prompted in a zero-shot manner using a natural language instruction. Specifically, we use the 8-bit quantized variants of \textit{Llama3.1-8B} \citep{dubey2024llama} and \textit{Phi3.5-mini} \citep{abdin2024phi}. We focus on SIB-200 for this baseline, as RoSPrompt is not designed for high-resource languages where smaller models cannot compete with large LLMs trained on extensive data. For transparency, results on MTOP and MLSUM are included in Appendix \ref{app:zero-shot-llm-baseline}, along with further details on this baseline.

\vspace{0.1cm}

\subsection{Technical Details}
Our experimental setup includes freezing all model parameters and appending a soft prompt to the initial input, as detailed in Section \ref{sec:related}. We start by initializing the soft prompt with the embeddings of the natural language prompt from \citet{zhao_pre-trained_2023}: "In this sentence, the topic is about". We then fine-tune this prompt using 8 randomly selected English samples from each class in DBPedia. Our methodology includes using translations of the original English label tokens into a diverse range of languages\footnote{We used the following languages as they are spoken by at least one member of our team: de, en, es, fa, fr, hi, ro, sv, uk, zh.}, and selecting words that tokenize as a single token for our multilingual label tokens.
We then assess the model's performance using the trained soft prompt on all three evaluation datasets across all supported languages. During evaluation, only the original English class names are needed, with no need for further translation efforts.

To account for variability in few-shot experiments, we repeat each experiment four times using different random seeds and report the average results.
\section{Results} \label{sec:results}
For each of the three models, our experimental findings are presented in Table \ref{tab:mtop_and_mlsum} for \texttt{MLSUM} and \texttt{MTOP}, across all languages. Given the extensive range of languages in SIB-200, we present average results for each major language family in Table \ref{tab:sib-200}, while detailed results for individual languages are shown in Appendix \ref{app:sib-200_full} (see Table \ref{tab:sib-200-all}).
Overall, our methodology shows a significant advantage over NPPrompt in nearly all cases. 
In particular, our training method, which leverages a mere 8 samples per class from an existing topic classification dataset, generates a soft prompt that is more effective for ZSC than a natural language prompt, demonstrating robust generalization capabilities for unseen classes.

Additionally, we observe that while larger generative LLMs slightly outperform the smaller RoSprompt-enhanced LLMs on high-resource languages, they significantly underperform, often worse than the random baseline, on low-resource languages, highlighting the effectiveness of our method in such scenarios.

\begin{table*}[t!]
\centering
    \setlength{\tabcolsep}{4pt}
    \renewcommand{\arraystretch}{1}
\begin{tabular}{c|c|cccccc||cccc}
 & & \multicolumn{6}{|c||}{\textbf{MTOP}} & \multicolumn{4}{c}{\textbf{MLSUM}} \\
\textbf{Model} &  & \textbf{de} & \textbf{en} & \textbf{es} & \textbf{fr} & \textbf{hi} & \textbf{th} & \textbf{de} & \textbf{es} & \textbf{fr} & \textbf{ru} \\ \hline
\multirow{4}{*}{\rotatebox[origin=c]{90}{\textbf{XGLM}}} & \textbf{RoSPrompt} & \textbf{54.99} & \textbf{64.31} & \textbf{58.95} & \textbf{55.38} & \textbf{56.47} & 47.59 & \textbf{79.47} & \textbf{70.77} & \textbf{71.60} & \textbf{62.66}\\
 & NPPrompt & 48.72 & 55.02 & 47.77 & 47.57 & 52.49 & \textbf{49.26} & 56.30 & 48.83 & 43.92 & 42.97 \\
 & NPPrompt-t & 26.63 & 55.02 & 42.03 & 19.14 & - & 33.27 & 61.22 & 21.68 & 31.97 & 38.24 \\
 & SPT & 30.52 & 31.98 & 32.51 & 30.98 & 28.69 & 29.46 & 63.12 & 53.26 & 54.00 & 53.68 \\ \hline
\multirow{4}{*}{\rotatebox[origin=c]{90}{\textbf{mT0}}} & \textbf{RoSPrompt} & \textbf{47.65} & \textbf{53.23} & \textbf{51.48} & \textbf{48.21} & \textbf{49.42} & \textbf{46.28} & \textbf{65.24} & 50.58 & \textbf{48.00} & 45.22 \\
 & NPPrompt & 43.14 & 46.35 & 48.57 & 43.60 & 46.04 & 38.37 & 65.07 & 48.10 & 43.23 & 43.14 \\ 
 & NPPrompt-t & 33.14 & 46.35 & 33.36 & 7.02 & - & 39.89 & 59.51 & 43.36 & 31.33 & 26.80 \\
 & SPT & 46.22 & 52.31 & 47.87 & 44.19 & 44.42 & 42.98 & 64.64 & \textbf{52.81} & 47.32 & \textbf{45.92} \\ \hline
\multirow{4}{*}{\rotatebox[origin=c]{90}{\textbf{XLM-R}}} & \textbf{RoSPrompt} & \textbf{55.64} & \textbf{63.93} & \textbf{54.79} & \textbf{52.91} & \textbf{62.25} & \textbf{53.28} & \textbf{81.77} & \textbf{65.46} & \textbf{60.66} & 53.39 \\
 & NPPrompt & 36.38 & 46.03 & 35.76 & 34.95 & 47.69 & 39.02 & 62.38 & 50.77 & 52.79 & \textbf{58.17} \\ 
 & NPPrompt-t & 35.25 & 46.03 & 35.29 & 28.47 & - & 47.05 & 72.95 & 41.89 & 38.18 & 48.37 \\
 & SPT & 39.10 & 43.75 & 35.29 & 36.35 & 40.04 & 37.55 & 69.00 & 57.83 & 50.40 & 49.59 \\ \hline
\end{tabular}
\caption{Comparison of accuracy scores on the \textbf{MTOP} and \textbf{MLSUM} datasets between RoSPrompt and baselines.}
\label{tab:mtop_and_mlsum}
\end{table*}

\begin{table*}[t!]
\centering
\setlength{\tabcolsep}{4pt}
\renewcommand{\arraystretch}{1}
\begin{tabular}{c|c|cccccccc}
\textbf{Model} &  & \textbf{\begin{tabular}[c]{@{}c@{}}Afro-\\ Asiatic\end{tabular}} & \textbf{\begin{tabular}[c]{@{}c@{}}Atlantic-\\ Congo\end{tabular}} & \textbf{\begin{tabular}[c]{@{}c@{}}Austro-\\ nesian\end{tabular}} & \textbf{\begin{tabular}[c]{@{}c@{}}Dravi-\\ dian\end{tabular}} & \textbf{\begin{tabular}[c]{@{}c@{}}Indo-\\ European\end{tabular}} & \textbf{\begin{tabular}[c]{@{}c@{}}Sino-\\ Tibetan\end{tabular}} & \textbf{Turkic} & \textbf{Uralic} \\ \hline
\multirow{4}{*}{\rotatebox[origin=c]{90}{\textbf{XGLM}}} & \textbf{RoSPrompt} & \textbf{69.12} & \textbf{65.32} & \textbf{73.04} & \textbf{64.95} & \textbf{70.80} & \textbf{72.92} & \textbf{72.55} & \textbf{71.51} \\
 & NPPrompt & 60.78 & 61.76 & 59.31 & 58.09 & 61.48 & 58.83 & 62.25 & 62.26 \\ 
 & NPPrompt-t & 53.92 & 58.82 & 63.73 & 58.09 & 54.41 & 53.68 & 62.25 & 40.69 \\
 & SPT & 59.19 & 55.51 & 66.05 & 58.15 & 60.94 & 54.05 & 60.42 & 66.54 \\ \hline
\multirow{4}{*}{\rotatebox[origin=c]{90}{\textbf{mT0}}} & \textbf{RoSPrompt} & \textbf{71.69} & \textbf{71.69} & \textbf{75.61} & \textbf{75.61} & \textbf{75.75} & \textbf{74.27} & \textbf{74.39} & \textbf{73.10} \\
 & NPPrompt & 57.11 & 59.13 & 59.95 & 61.64 & 61.52 & 62.42 & 61.03 & 63.40 \\ 
 & NPPrompt-t & 46.41 & 51.16 & 51.84 & 61.64 & 54.84 & 59.47 & 61.03 & 53.27 \\
 & SPT & 65.05 & 66.42 & 67.37 & 70.07 & 69.91 & 72.18 & 68.28 & 69.40 \\ \hline
\multirow{4}{*}{\rotatebox[origin=c]{90}{\textbf{XLM-R}}} & \textbf{RoSPrompt} & \textbf{72.67} & \textbf{65.69} & \textbf{71.69} & \textbf{66.91} & \textbf{68.65} & \textbf{68.63} & \textbf{67.89} & \textbf{70.59} \\
 & NPPrompt & 57.43 & 56.62 & 63.14 & 64.83 & 64.20 & 63.73 & 65.13 & 65.69 \\ 
 & NPPrompt-t & 45.26 & 38.24 & 57.25 & 64.83 & 52.05 & 57.84 & 65.13 & 57.03 \\
 & SPT & 56.78 & 52.33 & 61.96 & 64.49 & 61.65 & 65.28 & 61.40 & 57.31 \\ \hline
\multicolumn{2}{c|}{\textbf{Llama3.1-8B}} & 25.42 & 18.44 & 26.42 & 8.58 & 39.84 & 35.29 & 26.82 & 44.61 \\
\multicolumn{2}{c|}{\textbf{Phi-3.5-mini}} & 42.30 & 38.11 & 57.95 & 7.72 & 55.17 & 54.09 & 46.08 & 65.03 \\ \hline
\end{tabular}
\caption{Comparison of accuracy scores on the \textbf{SIB-200} dataset between RoSPrompt and baselines.}
\label{tab:sib-200}
\end{table*}


\section{Ablation Study}
To illustrate the individual contributions of each component in our training method, we carry out an ablation study. We assess the efficacy of our original method against variants lacking the loss penalty, contrastive label smoothing, and/or multilingual labels.

The outcomes of this study, presented in Table \ref{tab:mtop_ablation} for \texttt{MTOP} across three models, indicate that all three elements are integral to our method's success. Notably, the removal of the loss penalty leads to the most significant decline in performance for XGLM and mT0, while the lack of multilingual labels has the greatest negative impact on XLM-R. 


\begin{table}[H]
    \centering
    \setlength{\tabcolsep}{4pt}
    \renewcommand{\arraystretch}{1}
    \begin{tabular}{l|ccc}
        \textbf{} & \textbf{XGLM} & \textbf{mT0} & \textbf{XLM-R} \\ \hline
        \textbf{\textbf{RoSPrompt}} & \textbf{56.28} & 49.38 & \textbf{57.13} \\
        $\hspace{0.1cm}$ w/o penalty & 30.22 & 31.91 & 51.59 \\
        $\hspace{0.1cm}$ w/o LS & 50.05 & 49.71 & 48.18 \\
        $\hspace{0.1cm}$ w/o penalty \& LS & 29.38 & 41.53 & 51.26 \\
        $\hspace{0.1cm}$ w/o ML labels & 50.05 & \textbf{50.37} & 47.40 \\ \hline
    \end{tabular}
    \caption{Ablation study results for MTOP.}
    \label{tab:mtop_ablation}
\end{table}

This could potentially be attributed to XLM-R's enhanced code-switching capabilities \citep{winata_are_2021, zhang_multilingual_2023}, making it more efficient at using multilingual label tokens during training compared to XGLM and mT0.

\section{Generalized Zero-Shot Learning} \label{sec:GZSL}

In our initial experiments, training (\textit{seen}) and evaluation (\textit{unseen}) classes were distinct with merely minimal overlap. In contrast, the \textit{Generalized Zero-Shot Learning} (GZSL) settings, which mirror real-world situations more closely, involve evaluating on a mix of both seen and unseen classes. Models in this setting often struggle with overfitting to seen classes and fail to perform well on unseen classes \citep{xian_zero-shot_2019}. 

Therefore, we aim to investigate whether our method is also efficient under GZSL settings. For this, we fine-tune the soft prompt on a subset of classes from a dataset, then test it on the entire set of classes. Considering the potential variability resulting from the specific choice of seen and unseen classes, we repeat this process four times for each dataset and model, each time with a different subset of seen classes. We then average the F1 scores for seen and unseen classes and present them in Table \ref{tab:GZSL}. These experiments are conducted with all three models, but only for the SIB-200\footnote{For computational efficiency during this experiment, we limited our evaluation to a subset of ten linguistically diverse languages (en, ru, zh, de, ar, bn, ta, ko, my, sw) instead of all supported ones.} and MTOP datasets, as MLSUM does not support English, and has varying categories across languages.

\begin{table}[H]
    \centering
    \setlength{\tabcolsep}{2.5pt}
    \renewcommand{\arraystretch}{1.2}
    \begin{tabular}{p{0.5cm}|c|cc|cc}
         &  & \multicolumn{2}{c|}{\textbf{SIB-200}} & \multicolumn{2}{c}{\textbf{MTOP}} \\
         &  & \textit{Unseen} & \textit{Seen} & \textit{Unseen} & \textit{Seen} \\ \hline
        \multirow{2}{*}{\rotatebox{90}{\hspace{-0.15cm}\small{\textbf{XGLM}}}} & SPT & 20.02 & 48.56 & 28.04 & 49.04 \\
         & RoSPrompt & 48.68 & 49.60 & 62.41 & 61.50 \\ \hline
        \multirow{2}{*}{\rotatebox{90}{\hspace{-0.155cm}\small{\textbf{mT0}}}} & SPT & 31.32 & 39.44 & 32.26 & 23.49 \\
         & RoSPrompt & 67.11 & 65.44 & 39.33 & 52.98 \\ \hline
        \multirow{2}{*}{\rotatebox{90}{\hspace{-0.15cm}\small{\textbf{XLM-R}}}} & SPT & 26.88 & 53.78 & 17.63 & 43.88 \\
         & RoSPrompt & 56.64 & 55.68 & 62.54 & 57.96 \\ \hline
    \end{tabular}
    \caption{Comparison of average F1 scores for seen and unseen classes using standard SPT versus RoSPrompt.}
    \label{tab:GZSL}
\end{table}

For conventional SPT, there is a notable imbalance in performance between seen and unseen classes, with seen classes showing higher performance, suggesting overfitting to seen classes and poor generalization to unseen classes. However, when training the soft prompts using our method, the performance is more balanced, indicating improved generalization to unseen classes.

\section{Contextualizing Our Approach}
In this study, we acknowledge that comparing our approach with NPPrompt may not constitute an entirely fair comparison. RoSPrompt uses a small dataset for training, while NPPrompt directly leverages a PLM without additional fine-tuning. However, it is important to emphasize that the intent of our research is not to demonstrate RoSPrompt's performance superiority over NPPrompt. Instead, our objective is to illustrate how RoSPrompt's methodology can effectively improve cross-lingual transfer capabilities of natural language prompts. This aspect is vital as our findings indicate that merely converting hard prompts to soft prompts and then fine-tuning them using the standard SPT approach results in non-robust prompts which are ineffective for Generalized ZSC.

Additionally, while our paper focuses on topic classification, we believe that our approach could be equally effective for other types of classification tasks as well. Nonetheless, we emphasize the significance of zero-shot learning in topic classification, where classes often change more frequently over time or across domains, unlike in more stable tasks like sentiment analysis, where classes show less variation.

Furthermore, we want to emphasize the threefold efficiency of our approach: \textbf{1)} it is data efficient, requiring only a small number of labeled training samples from any comparable classification task; \textbf{2)} it is computationally efficient as fewer than 0.1\% of parameters are fine-tuned compared to full-model fine-tuning, reducing training time by approximately 50\% in our experiments; \textbf{3)} it is memory-efficient, as for $n$ training processes, besides the resulting $n$ prompts that take up a few hundred KBs at most, only one model copy is stored, in contrast to full-model fine-tuning where each model occupies several GBs of storage.

Moreover, while our method is theoretically applicable to larger models with billions of parameters, our primary target is smaller LLMs, which are often sufficient for tasks like zero-shot classification but need more focused guidance. These smaller multilingual models also excel in low-resource languages, where larger English-centric models, as we demonstrate, are less effective.

\section{Conclusion}
In this paper, we introduced \texttt{RoSPrompt}, a novel approach for cross-lingual zero-shot topic classification. It combines the advantages of few-shot SPT with the extensive knowledge acquired by language models in their pre-training phase. Our training method is designed for computational efficiency and incorporates three key components to enhance the standard SPT methodology, contributing to RoSPrompt's cross-lingual abilities and resilience to data distribution shifts.
\section*{Limitations}
Our research was conducted on datasets encompassing a variety of classes and data distributions. However, the absence of multilingual datasets across entirely distinct domains limits our ability to test the method's effectiveness in distant or niche domains. Therefore, while our results are promising within the domains we studied, they may not fully represent the model's capabilities across all specific domains.

In addressing the few-shot learning nature of our approach, varied the training samples across 4 iterations for each experiment to reduce potential biases. Nonetheless, the specific selection of these samples can still influence the outcomes due to the inherent characteristics of few-shot learning. This limitation suggests that our findings could be partially influenced by the particular datasets used, and might not entirely reflect the model's performance with different or broader data samples.
\section*{Ethics Statement}
In our work, we prioritized two key ethical aspects, through which we strive to contribute to the inclusive and responsible advancement of NLP technology.

\vspace{0.1cm}

\noindent \textbf{Language Diversity and Equity.} Our method aims to balance performance across various languages, addressing the common disparity in model effectiveness between high- and low-resource languages. By enhancing multilingual capabilities, \texttt{RoSPrompt} contributes towards more balanced performance across languages, ensuring fair and inclusive technology across diverse linguistic groups.

\vspace{0.1cm}

\noindent \textbf{Environmental Responsibility.} Our method is designed for computational efficiency, requiring fine-tuning of fewer than 1\% of parameters compared to traditional methods. This approach not only conserves computational resources but also aligns with environmental sustainability goals by reducing the energy consumption and carbon footprint associated with training and deploying NLP models.

\section*{Acknowledgment}
The author Cedric Lothritz is supported by the Luxembourg National Research Fund (FNR) PEARL program, grant agreement 16544475.

\bibliography{zotero, custom_lib}

\appendix

\section{Technical Details} \label{app:technical_details}
Access to the code used in our research will provided in the camer-ready version.

\subsection{Training}
We conducted all of our experiments using the Transformers library \citep{wolf_transformers_2020} and ran them on 4 A100 Nvidia GPUs within a few hours. We used AdamW \citep{loshchilov_decoupled_2019} as an optimizer. We provide the hyperparameters used during our experiments in Table \ref{tab:hyperparams}. Due to computational constraints, we did not perform exhaustive hyper-parameter optimization, but instead selected hyper-parameters that demonstrated satisfactory performance in preliminary experiments.

\begin{table}[h!]
    \centering
    \begin{tabular}{|c|c|c|c|}
    \hline
         & XGLM & XLM-R & mT0 \\ \hline
        Batch size & 8 & 8 & 8 \\ \hline
        Learning rate & 0.01 & 0.01 & 0.3 \\ \hline
        Epochs & 10 & 10 & 10\\ \hline
        $\alpha$ & 100 & 10 & 200 \\ \hline
        $\epsilon$ & 0.2 & 0.1 & 0.8 \\ \hline
        Prompt length & 8 & 8 & 9 \\ \hline
    \end{tabular}
    \caption{Hyperparameters}
    \label{tab:hyperparams}
\end{table}

\subsection{Evaluation}
During evaluation, NPPrompt \citep{zhao_pre-trained_2023} requires a parameter $k$, which is referred to as the \textit{neighborhood number}. In our experimental setup, for each model and dataset type, we selected the value of \textit{k} that achieved the highest average performance across the development sets of all supported languages. The specific values selected for $k$ in the evaluation of RoSPrompt, NPPrompt (including NPPrompt-t) and SPT are presented in Tables \ref{tab:k_ours}, \ref{tab:k_zhao} and \ref{tab:k_SPT} respectively.

\begin{table}[h!]
    \centering
    \begin{tabular}{|c|c|c|c|}
    \hline
         & XGLM & XLM-R & mT0 \\ \hline
        SIB-200 & 3 & 4 & 14 \\ \hline
        MTOP & 4 & 2 & 8 \\ \hline
        MLSUM & 300 & 5 & 7 \\ \hline
    \end{tabular}
    \caption{Chosen \textit{neighborhood number} $k$ values for \textbf{RoSPrompt}.}
    \label{tab:k_ours}
\end{table}

\begin{table}[h!]
    \centering
    \begin{tabular}{|c|c|c|c|}
    \hline
         & XGLM & XLM-R & mT0 \\ \hline
        SIB-200 & 4 & 3 & 6 \\ \hline
        MTOP & 3 & 2 & 5 \\ \hline
        MLSUM & 5 & 4 & 6 \\ \hline
    \end{tabular}
    \caption{Chosen \textit{neighborhood number} $k$ values for \textbf{NPPrompt} \citet{zhao_pre-trained_2023} and \textbf{NPPrompt-t} (\citet{zhao_pre-trained_2023} with translated hard prompt).}
    \label{tab:k_zhao}
\end{table}

\begin{table}[h!]
    \centering
    \begin{tabular}{|c|c|c|c|}
    \hline
         & XGLM & XLM-R & mT0 \\ \hline
        SIB-200 & 2 & 17 & 5 \\ \hline
        MTOP & 100 & 7 & 12 \\ \hline
        MLSUM & 200 & 16 & 7 \\ \hline
    \end{tabular}
    \caption{Chosen \textit{neighborhood number} $k$ values for \textbf{SPT}.}
    \label{tab:k_SPT}
\end{table}

\paragraph{Impact of Hyperparameters} 
RoSPrompt's training methodology primarily relies on two numerical hyperparameters: the contrastive label smoothing factor, denoted as $\epsilon$, and the penalty strength, represented by $\alpha$. 

\begin{figure}[H]
    \centering
    \includegraphics[width=\linewidth]{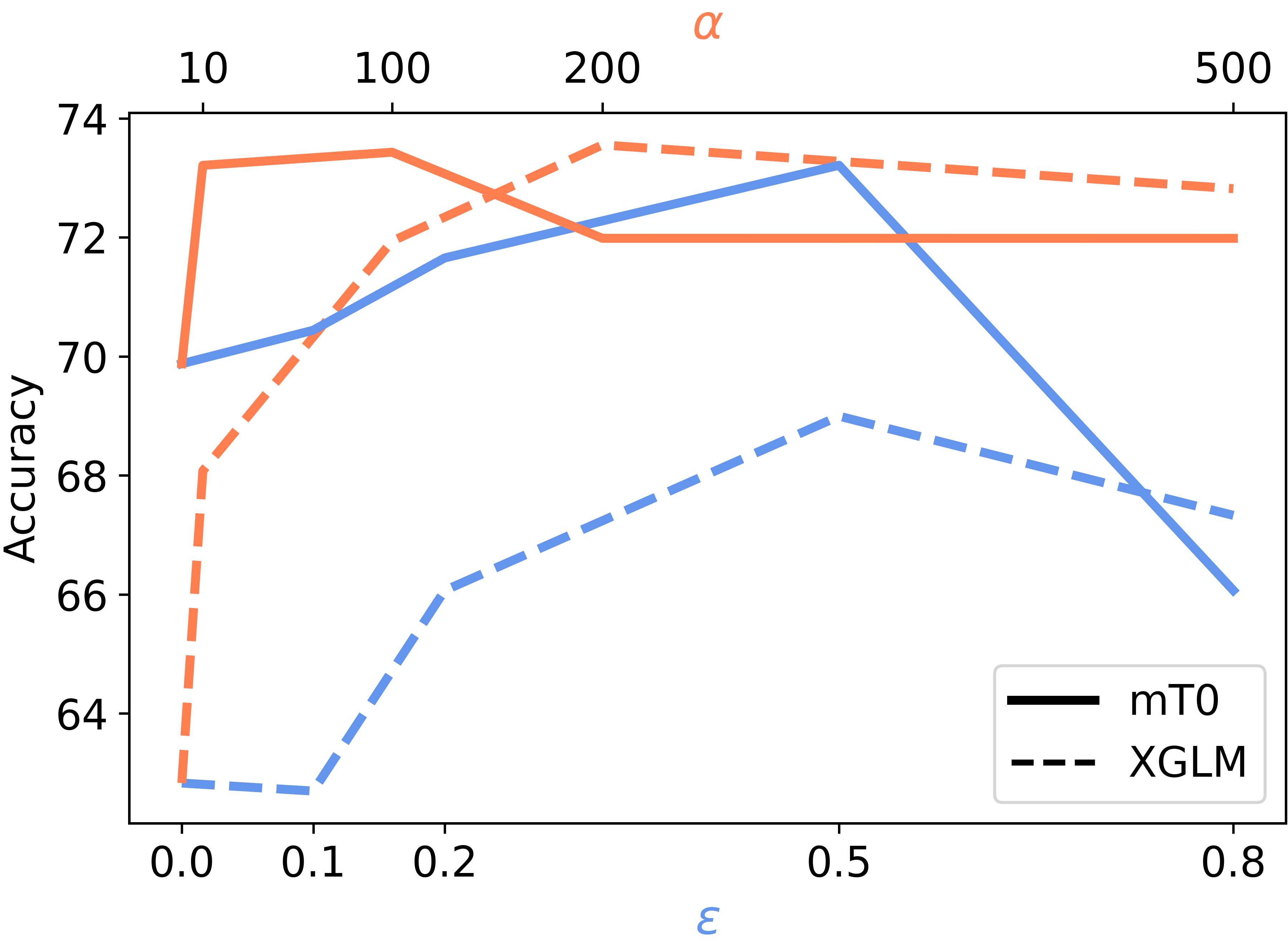}
    \caption{Average performance (accuracy) of RoSPrompt across 10 languages on SIB-200 for different values of {\color{NavyBlue}$\epsilon$} and {\color{RedOrange}$\alpha$}.}
    \label{fig:hyperparameters}
\end{figure}

In Figure \ref{fig:hyperparameters}, we illustrate RoSPrompt's performance using XGLM and mT0 on the SIB-200 dataset, using a diverse subset of languages\footnote{en, ru, zh, de, ar, bn, ta, ko, my, sw}, across various values for $\alpha$ and $\epsilon$, while maintaining the other hyperparameter at zero each time. Generally, we find that both excessively low and high values for $\alpha$ and $\epsilon$ do not lead to optimal outcomes.

\subsection{Datasets}
As source data to train the soft prompts, we use the \textbf{\texttt{DBPedia14}} ontology classification dataset\footnote{\url{https://huggingface.co/datasets/dbpedia_14}} \citep{lehmann_dbpedia_2015}. It is a subset of the English version of DBpedia 2014\footnote{\url{https://downloads.dbpedia.org/wiki-archive/data-set-2014.html}}, consisting of randomly chosen 560\,000 training and 70\,000 test samples equally distributed across 14 distinct classes. These classes represent the most common infobox categories on Wikipedia, including categories like \texttt{Company}, \texttt{Artist}, \texttt{Athlete}, \texttt{Village}, \texttt{Animal}, among others.

\vspace{5mm}

For evaluation we use three different multlingual datasets:

\noindent \textbf{\texttt{MLSUM}} \citep{scialom_mlsum_2020}, a multilingual news summarization dataset. However, each article-summary pair is also labeled with its respective news category. Therefore, in our experiments, we use, for each article, the summary for its classification. Given the differing data sources for different languages, the categories across languages slightly differ. More specifically we use articles on \texttt{society}, \texttt{politics}, \texttt{culture}, \texttt{sports}, \texttt{economy} and \texttt{science} for Spanish, Russian and French and articles on \texttt{politics}, \texttt{sports}, \texttt{economy}, \texttt{travel}, \texttt{car} and \texttt{education} for German. This selection amounts to 8935, 612, 5950, 5315 test samples for German, Russian, French and Spanish respectively. MLSUM is licensed under the MIT License\footnote{\url{https://opensource.org/license/mit/}}.

\vspace{5mm}

\noindent \textbf{\texttt{MTOP}}\footnote{\url{https://huggingface.co/datasets/mteb/mtop_domain}} \citep{li_mtop_2021}, a multilingual utterance classification dataset, featuring 11 different domains, such as \texttt{alarm}, \texttt{reminder}, \texttt{recipes} or \texttt{weather}. The dataset covers 6 languages: English, German, Spanish, French, Hindi and Thai, with respective test sample counts of 4386, 3549, 2998, 3193, 2789, and 2765. \texttt{MTOP} is licensed under the Creative Commons Attribution-ShareAlike 4.0 International License\footnote{\url{https://creativecommons.org/licenses/by-sa/4.0/}}.

\vspace{5mm}

\noindent \textbf{\texttt{SIB-200}}\footnote{\url{https://github.com/dadelani/sib-200}} \citep{adelani_sib-200_2024}, a multilingual topic classification dataset covering 203 languages. The dataset is derived from the FLORES-200 benchmark \citep{nllb_team_no_2022} and consists of 701 training, 99 validation and 204 test samples in each language. It features 7 distinct classes: \texttt{geography}, \texttt{politics}, \texttt{science/technology}, \texttt{travel}, \texttt{sports}, \texttt{health} and \texttt{entertainment}. \texttt{SIB-200} is licensed under the Apache License 2.0\footnote{\url{https://www.apache.org/licenses/LICENSE-2.0.txt}}.

\subsection{Models}
In our work, we use the following models:

\noindent $\textbf{\texttt{XGLM}}_\textit{564M}$\footnote{\url{https://huggingface.co/facebook/xglm-564M}} \citep{lin_few-shot_2022} is a decoder-only multilingual model supporting a diverse selection of 30 languages. Pre-trained on the CC100-XL dataset, an expansion of CC100 \citep{conneau_unsupervised_2020,wenzek_ccnet_2020}, it features 564 million parameters, 24 layers, a hidden dimension size of 1024, and 16 attention heads.

\vspace{5mm}

\noindent $\textbf{\texttt{XLM-R}}_\textit{Large}$ \footnote{\url{https://huggingface.co/xlm-roberta-large}} \citep{conneau_unsupervised_2020} is an encoder-only multilingual RoBERTa-based \citep{liu_roberta_2019} model supporting 100 languages, pre-trained on CC100 \citep{conneau_unsupervised_2020,wenzek_ccnet_2020} using the MLM objective. It consists of 550 million parameters, 24 hidden layers, a dimension of 1024, and 16 attention heads.

\vspace{5mm}

\noindent $\textbf{\texttt{mT0}}_\textit{Base}$\footnote{\url{https://huggingface.co/bigscience/mt0-base}}  \citep{muennighoff_crosslingual_2023} is an encoder-decoder model supporting 101 languages. It is an mT5 model \citep{xue_mt5_2021} that has been multi-task fine-tuned on the xP3 dataset\footnote{\url{https://huggingface.co/datasets/bigscience/xP3}} \citep{muennighoff_crosslingual_2023}.  It features 584 million parameters, 12 encoder and decoder layers, 12 attention heads, and a hidden dimension size of 768.

\begin{table*}[t!]
\centering
    \setlength{\tabcolsep}{4pt}
    \renewcommand{\arraystretch}{1.1}
\begin{tabular}{c|cccccc||cccc}
 & \multicolumn{6}{|c||}{\textbf{MTOP}} & \multicolumn{4}{c}{\textbf{MLSUM}} \\
\textbf{Model} & \textbf{de} & \textbf{en} & \textbf{es} & \textbf{fr} & \textbf{hi} & \textbf{th} & \textbf{de} & \textbf{es} & \textbf{fr} & \textbf{ru} \\ \hline
\textbf{Llama3.1-8B} & 83.26 & 93.50 & 85.32 & 83.15 & 84.69 & 75.26 & 78.13 & 70.21 & 68.27 & 53.92 \\
\textbf{Phi-3.5-mini} & 79.57 & 86.34 & 78.62 & 78.52 & 70.49 & 66.22 & 77.08 & 71.17 & 66.91 & 59.64 \\ \hline
\end{tabular}
\caption{Accuracy scores on the \textbf{MTOP} and \textbf{MLSUM} obtained through zero-shot prompting.}
\label{tab:mtop_and_mlsum_genLLM_baseline}
\end{table*}

\section{Additional Details on "Zero-Shot Prompting" Baseline} \label{app:zero-shot-llm-baseline}

For this baseline, we used the 8-bit quantized versions of \textit{Llama3.1-8B}\footnote{\url{https://huggingface.co/meta-llama/Llama-3.1-8B-Instruct}} \citep{dubey2024llama} and \textit{Phi-3.5-mini}\footnote{\url{https://huggingface.co/microsoft/Phi-3.5-mini-instruct}} \citep{abdin2024phi}, which have been designed with robust multilingual capabilities. \textit{Llama3.1-8B} is a transformer-based language model with 8.03 billion parameters, designed for efficient text generation tasks. \textit{Phi-3.5-mini}, a smaller variant, has 3.82 billion parameters and shares a similar transformer architecture optimized for lightweight inference. Both models were prompted using the prompt shown in Figure \ref{fig:LLM-baseline-prompt} and used with 8-bit quantization.

\begin{figure}[h!]
    \centering

        \begin{tcolorbox}[colback=gray!30, colframe=gray!70, sharp corners=all, boxrule=0.5mm, arc=5mm]
        \ttfamily
        
        I will provide text and potential categories, and I would like you to classify the text into one of the given categories based on its content. Please ensure the classification is accurate and consistent. \\
        
        Categories: \\
        - Label 1 \\
        - Label 2 \\
        - ...\\
        
        Text: "\{Document\}" \\
        
        Only return the category name.
    \end{tcolorbox}
    \caption{The prompt used for the \textbf{Zero-Shot LLMs} baseline with \textit{Llama3.1-8B} and \textit{Phi-3.5-mini}.}
    \label{fig:LLM-baseline-prompt}
\end{figure}

The results on MTOP and MLSUM are provided in Table \ref{tab:mtop_and_mlsum_genLLM_baseline}.

\section{Full Results for SIB-200} \label{app:sib-200_full}
Table \ref{tab:sib-200-all} presents the experimental results for each language on SIB-200, with average values per language family reported in Table \ref{tab:sib-200} in Section \ref{sec:results}.

{
    \small
    \setlength{\tabcolsep}{4pt}
    \onecolumn
    \begin{longtable}{c|cccc|cccc|cccc|cc}
              & \multicolumn{4}{c|}{\textbf{XGLM}}                                           & \multicolumn{4}{c|}{\textbf{mT0}}                                              & \multicolumn{4}{c|}{\textbf{XLM-R}}  &                                          \\ \hline
     \rule{0pt}{63pt} & \rotatebox{75}{\textbf{RoSPrompt}} & \rotatebox{75}{\textbf{NPPrompt}} & \rotatebox{75}{\textbf{NPPrompt-t}} & \rotatebox{75}{\textbf{SPT}} & \rotatebox{75}{\textbf{RoSPrompt}} & \rotatebox{75}{\textbf{NPPrompt}} & \rotatebox{75}{\textbf{NPPrompt-t}} & \rotatebox{75}{\textbf{SPT}} & \rotatebox{75}{\textbf{RoSPrompt}} & \rotatebox{75}{\textbf{NPPrompt}} & \rotatebox{75}{\textbf{NPPrompt-t}} & \rotatebox{75}{\textbf{SPT}} & \rotatebox{75}{\textbf{Llama3.1-8B}} &\rotatebox{75}{\textbf{Phi-3.5-mini}} \\ \hline
    \endfirsthead
    
    \hline
              & \multicolumn{4}{c|}{\textbf{XGLM}}                                           & \multicolumn{4}{c|}{\textbf{mT0}}                                              & \multicolumn{4}{c|}{\textbf{XLM-R}} & \\ \hline
     \rule{0pt}{63pt} & \rotatebox{75}{\textbf{RoSPrompt}} & \rotatebox{75}{\textbf{NPPrompt}} & \rotatebox{75}{\textbf{NPPrompt-t}} & \rotatebox{75}{\textbf{SPT}} & \rotatebox{75}{\textbf{RoSPrompt}} & \rotatebox{75}{\textbf{NPPrompt}} & \rotatebox{75}{\textbf{NPPrompt-t}} & \rotatebox{75}{\textbf{SPT}} & \rotatebox{75}{\textbf{RoSPrompt}} & \rotatebox{75}{\textbf{NPPrompt}} & \rotatebox{75}{\textbf{NPPrompt-t}} & \rotatebox{75}{\textbf{SPT}} &\rotatebox{75}{\textbf{Llama3.1-8B}} &\rotatebox{75}{\textbf{Phi-3.5-mini}} \\ \hline
    \endhead
    
    \endfoot
    
    \endlastfoot
    
    afr\_Latn &                      &       &       &       & 74.02                & 63.24 & 62.75 & \textbf{74.39}       & \textbf{69.85}       & 62.75          & 38.73 & 59.07                & 44.12          & {\ul \textbf{74.51}} \\
als\_Latn &                      &       &       &       & {\ul \textbf{75.37}} & 62.75 & 56.86 & 70.47                & \textbf{69.24}       & 65.69          & 24.02 & 61.15                & 30.88          & \textbf{58.82}       \\
amh\_Ethi &                      &       &       &       & \textbf{64.71}       & 55.39 & -     & 63.11                & {\ul \textbf{65.56}} & 59.80          & -     & 62.38                & 2.45           & \textbf{3.92}        \\
arb\_Arab & \textbf{69.12}       & 60.78 & 53.92 & 59.19 & \textbf{71.69}       & 62.25 & 64.71 & 71.08                & {\ul \textbf{72.67}} & 69.61          & 55.39 & 71.69                & 48.53          & \textbf{72.06}       \\
asm\_Beng &                      &       &       &       &                      &       &       &                      & {\ul \textbf{72.30}} & 62.75          & -     & 59.31                & \textbf{15.20} & 11.76                \\
azb\_Arab &                      &       &       &       & {\ul \textbf{65.07}} & 54.41 & -     & 58.58                & \textbf{59.80}       & 57.35          & -     & 49.02                & 23.04          & \textbf{37.25}       \\
azj\_Latn &                      &       &       &       & {\ul \textbf{75.37}} & 63.24 & -     & 69.73                & 67.52                & \textbf{69.61} & -     & 67.16                & 33.33          & \textbf{52.94}       \\
bel\_Cyrl &                      &       &       &       & {\ul \textbf{74.02}} & 62.75 & 60.29 & 69.36                & \textbf{67.89}       & 64.71          & 44.61 & 66.67                & 44.61          & \textbf{51.47}       \\
ben\_Beng & \textbf{69.36}       & 61.27 & 61.27 & 59.44 & {\ul \textbf{72.30}} & 57.35 & -     & 69.85                & \textbf{67.28}       & 64.22          & -     & 59.07                & \textbf{31.86}          & 25.00                \\
bos\_Latn &                      &       &       &       &                      &       &       &                      & {\ul \textbf{68.14}} & 64.71          & 60.29 & 58.70                & 48.53          & \textbf{60.29}       \\
bul\_Cyrl & \textbf{69.49}       & 63.73 & 57.35 & 63.73 & {\ul \textbf{77.45}} & 63.24 & 55.88 & 74.02                & \textbf{68.26}       & 66.18          & 59.80 & 67.77                & 51.96          & \textbf{67.65}       \\
cat\_Latn & \textbf{71.81}       & 65.69 & 59.80 & 59.19 & {\ul \textbf{79.90}} & 65.20 & 51.47 & 71.32                & \textbf{69.61}       & 64.71          & 59.31 & 67.77                & 53.43          & \textbf{75.98}       \\
ceb\_Latn &                      &       &       &       & {\ul \textbf{69.12}} & 61.27 & 54.41 & {\ul \textbf{69.12}} &                      &                &       &                      & 29.90          & \textbf{65.20}       \\
ces\_Latn &                      &       &       &       & {\ul \textbf{75.25}} & 62.25 & 50.98 & 71.81                & \textbf{65.93}       & 64.71          & 53.92 & 64.83                & 55.88          & \textbf{72.06}       \\
cym\_Latn &                      &       &       &       & \textbf{63.97}       & 55.88 & 32.84 & 63.48                & {\ul \textbf{65.69}} & 59.80          & 46.57 & 59.19                & 28.92          & \textbf{49.51}       \\
dan\_Latn &                      &       &       &       & \textbf{73.77}       & 64.22 & 61.76 & 72.06                & \textbf{69.00}       & 67.16          & 38.24 & 64.22                & 52.94          & {\ul \textbf{74.51}} \\
deu\_Latn & \textbf{70.22}       & 62.25 & 48.04 & 66.91 & \textbf{75.12}       & 66.18 & 62.25 & 72.55                & \textbf{69.36}       & 68.14          & 50.49 & 64.34                & 66.67          & {\ul \textbf{80.39}} \\
ell\_Grek & \textbf{69.36}       & 57.84 & 52.45 & 62.75 & {\ul \textbf{73.65}} & 58.82 & 65.20 & 69.49                & 69.12                & \textbf{70.10} & 47.06 & 64.83                & \textbf{51.47} & 43.63                \\
eng\_Latn & \textbf{73.53}       & 63.73 & 63.73 & 69.00 & \textbf{79.41}       & 65.20 & 65.20 & 73.53                & \textbf{68.01}       & 61.76          & 61.76 & 52.33                & 75.98          & {\ul \textbf{83.33}} \\
epo\_Latn &                      &       &       &       & {\ul \textbf{77.08}} & 67.16 & 40.20 & 73.04                & \textbf{67.65}       & 67.16          & 20.59 & 59.44                & 43.14          & \textbf{62.25}       \\
est\_Latn & \textbf{69.36}       & 61.76 & 22.55 & 65.56 & {\ul \textbf{73.28}} & 66.18 & 45.59 & 72.30                & \textbf{69.73}       & 66.18          & 47.06 & 58.21                & 36.76          & \textbf{55.88}       \\
eus\_Latn & \textbf{71.20}       & 63.73 & -     & 60.17 & {\ul \textbf{74.51}} & 63.73 & -     & 74.02                & \textbf{65.81}       & 58.82          & -     & 58.33                & 31.37          & \textbf{52.45}       \\
fin\_Latn & {\ul \textbf{73.65}} & 62.75 & 58.82 & 67.52 & \textbf{72.92}       & 62.25 & 52.45 & 67.28                & \textbf{71.45}       & 66.18          & 59.31 & 60.17                & 47.06          & \textbf{69.12}       \\
fra\_Latn & \textbf{71.08}       & 58.33 & 36.76 & 60.05 & \textbf{77.70}       & 64.22 & 58.33 & 70.71                & \textbf{66.18}       & 65.20          & 31.86 & 60.17                & 65.20          & {\ul \textbf{78.92}} \\
gaz\_Latn &                      &       &       &       &                      &       &       &                      & {\ul \textbf{44.24}} & 35.29          & 29.41 & 38.48                & 9.31           & \textbf{25.98}       \\
gla\_Latn &                      &       &       &       & {\ul \textbf{60.42}} & 48.53 & 38.24 & 56.13                & \textbf{59.19}       & 54.90          & 41.67 & 53.19                & 13.24          & \textbf{30.88}       \\
gle\_Latn &                      &       &       &       & 69.00                & 60.78 & 29.90 & {\ul \textbf{69.24}} & \textbf{63.60}       & 57.84          & 44.61 & 56.86                & 19.61          & \textbf{42.16}       \\
glg\_Latn &                      &       &       &       & \textbf{76.47}       & 67.65 & 47.55 & \textbf{76.47}       & \textbf{68.75}       & 64.71          & 52.45 & 68.50                & 52.94          & {\ul \textbf{77.94}} \\
guj\_Gujr &                      &       &       &       & {\ul \textbf{74.02}} & 65.20 & -     & 69.24                & \textbf{68.26}       & 60.29          & -     & 64.09                & \textbf{6.86}  & 2.94                 \\
hat\_Latn & \textbf{65.81}       & 59.80 & 62.25 & 48.41 & {\ul \textbf{69.73}} & 55.39 & 40.69 & 67.40                &                      &                &       &                      & 18.63          & \textbf{54.41}       \\
hau\_Latn &                      &       &       &       & 60.91                & 51.47 & 42.65 & {\ul \textbf{61.64}} & \textbf{60.54}       & 59.31          & 45.59 & 51.10                & 23.53          & \textbf{33.33}       \\
heb\_Hebr &                      &       &       &       & {\ul \textbf{73.41}} & 59.80 & 51.47 & 68.26                & \textbf{67.28}       & 65.20          & 41.67 & 64.83                & 52.45          & \textbf{58.33}       \\
hin\_Deva & \textbf{67.89}       & 62.25 & -     & 58.09 & {\ul \textbf{72.43}} & 62.25 & -     & 71.45                & \textbf{71.20}       & 65.69          & -     & 65.56                & 50.00          & \textbf{55.39}       \\
hrv\_Latn &                      &       &       &       &                      &       &       &                      & {\ul \textbf{68.26}} & 66.67          & 62.75 & 58.09                & 48.04          & \textbf{64.22}       \\
hun\_Latn &                      &       &       &       & {\ul \textbf{72.92}} & 61.76 & -     & 68.63                & \textbf{69.98}       & 64.71          & -     & 53.55                & 50.00          & \textbf{70.10}       \\
hye\_Armn &                      &       &       &       & {\ul \textbf{71.69}} & 61.27 & 65.69 & 66.54                & \textbf{70.10}       & 66.67          & 68.14 & 63.60                & 11.27          & \textbf{25.00}       \\
ibo\_Latn &                      &       &       &       & {\ul \textbf{71.45}} & 62.25 & 55.88 & 68.75                &                      &                &       &                      & 24.51          & \textbf{34.80}       \\
ind\_Latn & \textbf{73.04}       & 59.31 & 63.73 & 66.05 & \textbf{75.61}       & 67.16 & 57.84 & 72.92                & \textbf{71.69}       & 67.16          & 63.73 & 67.28                & 52.94          & {\ul \textbf{79.41}} \\
isl\_Latn &                      &       &       &       & {\ul \textbf{70.96}} & 61.27 & 53.92 & 69.61                & \textbf{67.77}       & 67.65          & 39.22 & 58.82                & 24.51          & \textbf{48.53}       \\
ita\_Latn & \textbf{72.43}       & 63.24 & 48.53 & 63.24 & \textbf{75.00}       & 63.24 & 53.43 & 73.16                & \textbf{66.79}       & 65.69          & 44.61 & 63.73                & 64.22          & {\ul \textbf{79.41}} \\
jav\_Latn &                      &       &       &       &                      &       &       &                      & 66.54                & 60.78          & 51.96 & {\ul \textbf{66.67}} & 19.12          & \textbf{61.76}       \\
jpn\_Jpan & \textbf{72.30}       & 62.25 & -     & 59.68 & {\ul \textbf{75.49}} & 62.75 & -     & 72.18                & \textbf{69.12}       & 64.22          & -     & 64.22                & 49.51          & \textbf{72.55}       \\
kan\_Knda &                      &       &       &       & {\ul \textbf{72.92}} & 62.25 & -     & 68.75                & 65.20                & \textbf{66.18} & -     & 66.05                & \textbf{8.82}  & 2.45                 \\
kat\_Geor &                      &       &       &       & {\ul \textbf{74.14}} & 62.25 & 58.33 & 72.30                & \textbf{70.47}       & 66.18          & 55.88 & 65.93                & 5.39           & \textbf{18.63}       \\
kaz\_Cyrl &                      &       &       &       & {\ul \textbf{76.96}} & 63.24 & -     & 71.20                & \textbf{73.04}       & 70.10          & -     & 69.24                & 28.92          & \textbf{56.86}       \\
khk\_Cyrl &                      &       &       &       & 69.73                & 58.33 & -     & {\ul \textbf{69.85}} & \textbf{65.69}       & 60.29          & -     & 53.92                & 20.10          & \textbf{34.80}       \\
khm\_Khmr &                      &       &       &       & {\ul \textbf{71.57}} & 65.69 & 61.27 & 70.71                & 66.54                & \textbf{66.67} & 52.45 & 64.71                & 3.43           & \textbf{4.90}        \\
kir\_Cyrl &                      &       &       &       & {\ul \textbf{70.83}} & 60.78 & -     & 69.00                & \textbf{69.49}       & 66.67          & -     & 61.76                & 22.55          & \textbf{50.00}       \\
kmr\_Latn &                      &       &       &       & 57.35                & 52.94 & -     & \textbf{57.84}       & {\ul \textbf{64.09}} & 59.31          & -     & 61.40                & 20.10          & \textbf{43.14}       \\
kor\_Hang & \textbf{68.38}       & 60.78 & -     & 62.99 & \textbf{71.57}       & 59.31 & -     & 68.87                & \textbf{69.00}       & 63.24          & -     & 62.62                & 45.59          & {\ul \textbf{73.53}} \\
lao\_Laoo &                      &       &       &       & {\ul \textbf{74.39}} & 65.69 & 61.76 & 74.02                & \textbf{68.63}       & 61.27          & 60.78 & 65.44                & \textbf{3.92}  & 2.94                 \\
lit\_Latn &                      &       &       &       & {\ul \textbf{74.02}} & 64.71 & 64.71 & 69.98                & 66.05                & \textbf{67.65} & 25.98 & 57.35                & 36.76          & \textbf{59.31}       \\
ltz\_Latn &                      &       &       &       & 66.42                & 55.88 & 44.61 & {\ul \textbf{66.54}} &                      &                &       &                      & 23.53          & \textbf{65.69}       \\
lvs\_Latn &                      &       &       &       & {\ul \textbf{74.26}} & 61.76 & 47.55 & 70.22                & \textbf{69.12}       & 62.75          & 24.51 & 48.53                & 38.24          & \textbf{56.37}       \\
mal\_Mlym &                      &       &       &       & {\ul \textbf{72.06}} & 58.33 & -     & 68.26                & \textbf{70.10}       & 66.18          & -     & 64.83                & 8.33           & \textbf{10.78}       \\
mar\_Deva &                      &       &       &       & {\ul \textbf{71.45}} & 61.27 & -     & 67.52                & \textbf{67.03}       & 61.27          & -     & 57.60                & \textbf{35.29}          & 33.33                \\
mkd\_Cyrl &                      &       &       &       & {\ul \textbf{76.47}} & 60.78 & 63.73 & 72.06                & \textbf{70.83}       & 61.27          & 59.80 & 62.38                & 40.20          & \textbf{63.24}       \\
mlt\_Latn &                      &       &       &       & {\ul \textbf{68.26}} & 58.82 & 27.45 & 66.18                &                      &                &       &                      & 28.92          & \textbf{65.69}       \\
mri\_Latn &                      &       &       &       & 56.13                & 47.06 & 26.47 & {\ul \textbf{58.09}} &                      &                &       &                      & 11.76          & \textbf{32.35}       \\
mya\_Mymr & {\ul \textbf{72.30}} & 62.75 & -     & 61.40 & \textbf{71.32}       & 58.33 & -     & 70.71                & \textbf{68.38}       & 61.76          & -     & \textbf{68.38}       & 2.45           & \textbf{3.92}        \\
nld\_Latn &                      &       &       &       & \textbf{76.35}       & 63.73 & 66.18 & 74.63                & \textbf{70.34}       & 68.14          & 51.96 & 60.29                & 58.82          & {\ul \textbf{79.41}} \\
nno\_Latn &                      &       &       &       & \textbf{74.14}       & 63.73 & -     & 69.24                & \textbf{69.98}       & 62.75          & -     & 66.30                & 40.69          & {\ul \textbf{75.00}} \\
nob\_Latn &                      &       &       &       & {\ul \textbf{75.25}} & 62.75 & 58.82 & 70.59                & \textbf{70.96}       & 64.71          & 38.73 & 62.62                & 51.47          & \textbf{73.53}       \\
npi\_Deva &                      &       &       &       & {\ul \textbf{73.04}} & 62.25 & -     & 70.34                & \textbf{69.85}       & 66.18          & -     & 65.44                & 25.98          & \textbf{45.10}       \\
nya\_Latn &                      &       &       &       & {\ul \textbf{70.59}} & 61.76 & -     & 69.24                &                      &                &       &                      & 15.69          & \textbf{38.24}       \\
pan\_Guru &                      &       &       &       & {\ul \textbf{72.30}} & 61.76 & -     & 71.94                &                      &                &       &                      & \textbf{9.31}           & 2.94                 \\
pbt\_Arab &                      &       &       &       & 66.91                & 55.88 & -     & {\ul \textbf{67.65}} & \textbf{65.07}       & 62.75          & -     & 62.62                & 23.04          & \textbf{38.24}       \\
pes\_Arab &                      &       &       &       & {\ul \textbf{75.00}} & 60.29 & -     & 68.75                & \textbf{68.75}       & 65.69          & -     & 61.03                & 45.10          & \textbf{52.94}       \\
plt\_Latn &                      &       &       &       & {\ul \textbf{67.28}} & 55.88 & -     & 66.54                & \textbf{62.99}       & 54.90          & -     & 48.04                & 13.24          & \textbf{42.65}       \\
pol\_Latn &                      &       &       &       & \textbf{75.98}       & 62.75 & 54.41 & 74.14                & \textbf{66.42}       & 64.71          & 47.55 & 64.95                & 57.84          & {\ul \textbf{78.43}} \\
por\_Latn & \textbf{72.92}       & 64.71 & 50.49 & 66.18 & \textbf{76.35}       & 67.16 & 37.25 & 74.02                & \textbf{70.34}       & 66.67          & 57.35 & 68.63                & 60.78          & {\ul \textbf{77.45}} \\
quy\_Latn & {\ul \textbf{45.71}} & 44.12 & -     & 32.48 &                      &       &       &                      &                      &                &       &                      & 14.71          & \textbf{41.18}       \\
ron\_Latn &                      &       &       &       & 72.43                & 64.22 & 37.25 & {\ul \textbf{75.86}} & \textbf{71.32}       & 68.63          & 56.86 & 62.01                & 50.00          & \textbf{72.55}       \\
rus\_Cyrl & \textbf{70.22}       & 60.78 & 57.84 & 64.95 & \textbf{75.49}       & 63.73 & 61.27 & 72.55                & 68.75                & \textbf{69.12} & 65.69 & 67.28                & 59.80          & {\ul \textbf{75.98}} \\
san\_Deva &                      &       &       &       &                      &       &       &                      & {\ul \textbf{64.34}} & 62.25          & -     & 59.93                & 18.63          & \textbf{41.18}       \\
sin\_Sinh &                      &       &       &       & {\ul \textbf{72.18}} & 60.29 & -     & 69.12                & \textbf{68.14}       & 62.25          & -     & 60.05                & \textbf{4.41}  & 3.92                 \\
slk\_Latn &                      &       &       &       & \textbf{70.96}       & 62.25 & 31.86 & 69.61                & 67.77                & \textbf{69.12} & 65.20 & 63.85                & 44.61          & {\ul \textbf{71.57}} \\
slv\_Latn &                      &       &       &       & 72.43                & 62.25 & 52.94 & {\ul \textbf{73.53}} & \textbf{65.93}       & 64.71          & 61.27 & 62.62                & 40.20          & \textbf{67.65}       \\
smo\_Latn &                      &       &       &       & 60.91                & 50.49 & 49.51 & {\ul \textbf{63.48}} &                      &                &       &                      & 13.24          & \textbf{33.33}       \\
sna\_Latn &                      &       &       &       & {\ul \textbf{67.03}} & 59.31 & 48.53 & 64.71                &                      &                &       &                      & 15.20          & \textbf{39.71}       \\
snd\_Arab &                      &       &       &       & \textbf{65.32}       & 58.33 & 43.63 & 63.85                & {\ul \textbf{67.40}} & 58.82          & 16.67 & 55.15                & 26.47          & \textbf{32.84}       \\
som\_Latn &                      &       &       &       & 59.44                & 54.90 & 36.76 & {\ul \textbf{60.05}} & \textbf{59.19}       & 55.39          & 39.71 & 52.21                & 12.75          & \textbf{36.76}       \\
sot\_Latn &                      &       &       &       & {\ul \textbf{70.34}} & 58.33 & 46.57 & 67.40                &                      &                &       &                      & 14.71          & \textbf{34.80}       \\
spa\_Latn & \textbf{74.39}       & 60.29 & 44.12 & 55.88 & \textbf{78.19}       & 67.65 & 56.86 & 74.02                & \textbf{66.67}       & 65.69          & 46.57 & 68.50                & 68.63          & {\ul \textbf{80.39}} \\
srp\_Cyrl &                      &       &       &       & {\ul \textbf{76.84}} & 64.22 & 64.71 & 71.08                & \textbf{69.36}       & 62.75          & 57.35 & 59.56                & 46.08          & \textbf{60.78}       \\
sun\_Latn &                      &       &       &       & {\ul \textbf{73.04}} & 60.29 & 54.90 & 70.83                & \textbf{68.14}       & 67.16          & 57.35 & 67.40                & 17.16          & \textbf{62.25}       \\
swe\_Latn &                      &       &       &       & \textbf{72.92}       & 62.25 & 55.88 & 70.96                & \textbf{71.81}       & 67.16          & 57.35 & 66.67                & 54.90          & {\ul \textbf{75.49}} \\
swh\_Latn & \textbf{65.32}       & 61.76 & 58.82 & 55.51 & {\ul \textbf{71.69}} & 61.76 & 49.51 & 68.87                & \textbf{65.69}       & 62.75          & 50.00 & 55.27                & 28.43          & \textbf{44.12}       \\
tam\_Taml & \textbf{67.65}       & 60.78 & -     & 61.40 & {\ul \textbf{76.10}} & 62.75 & -     & 71.32                & \textbf{66.05}       & 63.24          & -     & 63.48                & 10.29          & \textbf{15.20}       \\
tel\_Telu & \textbf{62.25}       & 55.39 & -     & 54.90 & {\ul \textbf{75.12}} & 63.24 & -     & 71.94                & \textbf{67.77}       & 63.73          & -     & 63.60                & \textbf{6.86}           & 2.45                 \\
tgk\_Cyrl &                      &       &       &       & {\ul \textbf{70.47}} & 60.29 & 57.84 & 66.79                &                      &                &       &                      & 23.04          & \textbf{35.78}       \\
tgl\_Latn &                      &       &       &       & {\ul \textbf{71.94}} & 63.73 & 59.80 & 68.63                &                      &                &       &                      & 41.67          & \textbf{70.10}       \\
tha\_Thai & \textbf{67.65}       & 58.82 & 53.92 & 59.31 & {\ul \textbf{73.28}} & 62.25 & 63.73 & 70.96                & \textbf{69.12}       & 65.69          & 54.90 & 68.87                & 52.45          & \textbf{54.90}       \\
tur\_Latn & \textbf{72.55}       & 62.25 & -     & 60.42 & {\ul \textbf{74.39}} & 61.27 & -     & 71.69                & \textbf{67.89}       & 65.20          & -     & 63.11                & 42.16          & \textbf{70.10}       \\
uig\_Arab &                      &       &       &       &                      &       &       &                      & \textbf{67.16}       & 62.25          & -     & 59.93                & \textbf{15.20} & 9.31                 \\
ukr\_Cyrl &                      &       &       &       & {\ul \textbf{73.65}} & 62.75 & 63.24 & 71.32                & 67.89                & \textbf{69.61} & 51.96 & 66.79                & 50.00          & \textbf{69.61}       \\
urd\_Arab & \textbf{67.65}       & 56.86 & -     & 55.27 & {\ul \textbf{71.81}} & 59.80 & -     & 69.00                & \textbf{70.83}       & 61.76          & -     & 65.69                & \textbf{53.92} & 40.69                \\
uzn\_Latn &                      &       &       &       & {\ul \textbf{74.63}} & 63.24 & -     & 69.49                & \textbf{69.00}       & 64.71          & -     & 59.56                & 22.55          & \textbf{46.08}       \\
vie\_Latn & \textbf{69.61}       & 66.18 & 58.33 & 59.31 & {\ul \textbf{72.79}} & 62.25 & 62.75 & 70.34                & 66.79                & \textbf{68.14} & 54.90 & 62.50                & 44.12          & \textbf{69.61}       \\
xho\_Latn &                      &       &       &       & {\ul \textbf{69.36}} & 61.27 & 50.49 & 67.89                & \textbf{52.82}       & 50.49          & 26.47 & 49.39                & 16.18          & \textbf{42.16}       \\
ydd\_Hebr &                      &       &       &       & \textbf{60.66}       & 53.43 & 43.14 & 59.80                & {\ul \textbf{62.25}} & 51.47          & 30.39 & 43.14                & 16.67          & \textbf{18.14}       \\
yor\_Latn &                      &       &       &       & 55.88                & 49.51 & 40.69 & {\ul \textbf{59.07}} &                      &                &       &                      & 14.22          & \textbf{30.88}       \\
zho\_Hans & \textbf{73.53}       & 54.90 & 44.61 & 46.69 & \textbf{77.21}       & 63.24 & 58.33 & 74.88                & \textbf{68.87}       & 63.24          & 56.86 & 65.56                & 54.90          & {\ul \textbf{78.92}} \\
zho\_Hant &                      &       &       &       & \textbf{74.14}       & 65.69 & 61.76 & 70.96                & \textbf{70.22}       & 66.18          & 54.90 & 61.89                & 48.53          & {\ul \textbf{79.41}} \\
zsm\_Latn &                      &       &       &       & \textbf{73.16}       & 65.69 & 55.88 & 69.36                & \textbf{69.61}       & 65.69          & 58.33 & 60.42                & 38.73          & {\ul \textbf{74.51}} \\
zul\_Latn &                      &       &       &       & {\ul \textbf{67.77}} & 58.82 & 55.88 & 65.44                &                      &                &       &                      & 18.63          & \textbf{40.20}       \\ \hline
    \caption{Comparison of accuracy scores on the SIB-200 dataset between RoSPrompt and different baselines across all supported languages. For each language, the best overall result is {\ul underlined}, and the best result within each column group is highlighted in \textbf{bold}.}\label{tab:sib-200-all}
    \end{longtable}
}
\end{document}